\documentclass[review]{elsarticle}
\usepackage{todonotes}
\usepackage{easylist}
\usepackage{amssymb}
\usepackage{amsmath}
\usepackage{subcaption}
\usepackage[breaklinks]{hyperref}
\usepackage{url}
\usepackage{textcomp}
\usepackage{verbatim}
\usepackage[ruled,vlined,linesnumbered]{algorithm2e}
\usepackage{graphicx}
\usepackage{pgfplots}
\pgfplotsset{compat=1.14}
\pgfplotsset{compat=newest}
\pgfplotsset{plot coordinates/math parser=false}
\usepackage{tikzscale}
\usetikzlibrary{matrix,chains,positioning,decorations.pathreplacing,arrows}
\usepackage{tikz-qtree,tikz-qtree-compat}
\usetikzlibrary{calc}
\journal{arXiv}

\begin{document}
\begin{frontmatter}

%\title{Disease module detection with a network-derived explainable decision forest}

\title{Graph-guided random forest for gene set selection}

%\title{Actionable Explainable AI for Systems Biology: \\ A network-derived decision forest for \\ disease module discovery}

%\title{Network Module Detection from Multi-Modal Node Features with a Greedy Decision Forest for Actionable Explainable AI}


\author[MUG]{Bastian Pfeifer \corref{mycorrespondingauthor}}
\cortext[mycorrespondingauthor]{Corresponding author}
\ead{bastian.pfeifer@medunigraz.at}
\author[WUT]{Hubert Baniecki}
\author[MUG]{Anna Saranti}
\author[WUT]{Przemyslaw Biecek}
\author[MUG,amii]{Andreas Holzinger} 

\address[MUG]{Institute for Medical Informatics Statistics and Documentation, Medical University Graz, Austria}
\address[amii]{Alberta Machine Intelligence Institute, Canada}
\address[WUT]{MI${}^{\,2}$DataLab, Faculty of Mathematics and Information Science, Warsaw University of Technology, Poland}

\begin{abstract} 
Machine learning methods can detect complex relationships between variables, but usually do not exploit domain knowledge. This is a limitation because in many scientific disciplines, such as systems biology, domain knowledge is available in the form of graphs or networks, and its use can improve model performance. We need network-based algorithms that are versatile and applicable in many research areas. 
In this work, we demonstrate subnetwork detection based on multi-modal node features using a novel Greedy Decision Forest with inherent interpretability. The latter will be a crucial factor to retain experts and gain their trust in such algorithms. To demonstrate a concrete application example, we focus on bioinformatics, systems biology and particularly biomedicine, but the presented methodology is applicable in many other domains as well. Systems biology is a good example of a field in which statistical data-driven machine learning enables the analysis of large amounts of multi-modal biomedical data. This is important to reach the future goal of precision medicine, where the complexity of patients is modeled on a system level to best tailor medical decisions, health practices and therapies to the individual patient. Our proposed approach can help to uncover disease-causing network modules from multi-omics data to better understand complex diseases such as cancer. 

\end{abstract}


\end{frontmatter}

%\linenumbers

%%%%%%%%%%%%%%%% (please use short comment symbols - as they are taking much space elsewhere

% this is just for us and is only required in certain journals
%\section*{Article Summary}
%What is already known on the topic to the international research community?
%\begin{itemize}
%    \item 1)
%    \item 2)
%    \item 3)
%\end{itemize}

%What this paper contributes to the international research community ?

%\begin{itemize}
%    \item 1)
%    \item 2)
%    \item 3)
%\end{itemize}
%%%%%%%%%%%%%%%%%%%

%%% HIGHLIGHTS
%\section*{\textbf{Highlights}}
%\begin{itemize}

%    \item We present a novel tree-based subnetwork detection algorithm 
%    \item Our approach supports multi-modal node features of arbitrary dimensions \item Moreover, our approach allows for multi-graphs as an input
    
%  \end{itemize}

%%%%%%%%%%%%%%%%%%%%
\section{Introduction}
\label{Introduction}

Network-based algorithms are utilized in most areas of research and industry. They find their application in virtually all areas, from agriculture to zoology, and are particularly useful in systems biology \cite{JurisicaEtAl:2015:NetworkBiology, DehmerEtAl:2016:BigDataComplexNetworks}. Networks are very important to solve problems in decision making and knowledge discovery \cite{KickertEtAl:1997:NetworksPolicy}. Such networks can represent many real-world phenomena and are technically described by graphs, which provide a unifying abstraction by which such real-world networks can be represented, explored, predicted and discovered \cite{Sherif:2021:graphs}. Most importantly, such structures help to make complex phenomena re-traceable, transparent, interpretable and thus explainable to human experts. Better interpretability promotes developer understanding. However, it also means easier understanding and better explanation of future predictions for end users, thus promoting reasonable and data-driven decisions to make personalized choices that can ultimately lead to higher quality healthcare services \cite{StiglicZitnik:2020:InterpretabilityHealth}.

What we need in the future are context-adaptive methods, i.e. systems that construct contextual explanatory models for classes of real-world phenomena. This is a goal of so-called explainable AI \cite{ArrietaHerrera:2020:exaiINFFUS}, which is actually not a new field; rather, the problem of explainability is as old as AI itself. While the rule-based approaches of early AI were comprehensible "glass box" approaches, at least in narrow domains, their weakness lay in dealing with the uncertainties of the real world \cite{Holzinger:2018:IEEE-DISA}. Actionable Explainable AI (AXAI) is intended to help promote trust-building features by bringing decision analytic perspectives and human domain knowledge directly into the AI pipeline. \cite{LinkovEtAl.2020:ActionableAI, HolzNat:2022:Inffus}.
 
In this work, we focus as example application on systems biology with a particular emphasis on biomedicine. Recent developments in bioinformatics and machine learning have made it possible to analyze huge amounts of biomedical data, paving the way for future precision medicine. A grand goal of precision medicine is in modeling the complexity of patients to tailor medical decisions, health practices and therapies to the individual patient. To reach this goal efficient computational methods, algorithms and tools are needed to discover knowledge and to interactively gain insight into the data \cite{Holzinger:2014:trends}. Network-based approaches help uncover disease-causing interactions between genes to better understand diseases such as cancer \cite{IdekerSharan:2008:ProteinNetworks, JeanquartierEtAl:2018:InSilicoCancer}. In systems biology, for instance, network analyses are performed for protein-protein co-expression networks as well as for metabolomics data through metabolomic pathway analyses.

Recent evidence suggests that complex diseases, such as cancer, need to be studied in a multi-modal feature space comprising multiple biological entities (multi-omics), as diverse components contribute to a single outcome \cite{HolzingerHaibeJurisica:2019:ImagingIntegration, HolzingerEtAl:2021:GraphFusion}. Multi-omics clustering methods are  successfully utilized to detecting disease subtypes, that is patient groups with similar molecular characteristics. Recently developed multi-omics clustering approaches include SNF \cite{wang2014similarity}, PINSplus \cite{Nguyen:2020:MultiviewLearningMultiomics}, and HC-fused \cite{pfeifer2021hierarchical}. 

Another typical research goal is the discovery of disease-causing genes to efficiently monitor disease progression. These so-called biomarkers are often detected from multi-modal omics data sets and here it is important to allow physicians to interactively intervene, read out corresponding patterns from the data and adjust their treatment accordingly. 

Feature selection (FS) algorithms are often used to detect such patterns. One popular family of FS algorithms is based on the random forest (RF) classifier. RFs can efficiently handle heterogeneous data types, don't need normalization of their values \cite{Zheng:2018:FeatureEngineering}, compute feature importances and are therefore particularly suited for the multi-modal case. 
One example of a popular RF-based feature selector is the Boruta algorithm \cite{kursa2010feature}. The heuristic introduced by the authors quantifies the importance of a feature by the loss of accuracy of classification caused by randomly permuted probes (\textit{shadow features}) of the original feature set. Recent work combines the Boruta algorithm with Shapley values \cite{eoghan_keany_2020_4247618}. Another RF-based feature selection algorithm is introduced in \cite{deng2013gene}. The authors have developed a regularized random forest approach that penalizes the selection of a new feature for splitting a node when its information gain is similar to the features used in previous splits.

However, most of these methods treat genes/features independently or do not account for any dependencies between variables. Contrarily, approaches which do account for dependencies between genes are tailored towards the detection of network modules with correlated features only (like in co-expression analyses). They typically utilize community detection algorithms \cite{choobdar2019assessment} or unsupervised clustering. Disease-causing genes, however, may function in a multivariate manner without necessarily being correlated. 

Most recently, the machine learning community has developed deep learning techniques which can be applied to networks comprising edge and node features of arbitrary dimensions. Graph neural networks (GNNs) \cite{Kipf:2016:SemiSupervisedClassification, Xu:2018:PowerfulGIN} are very versatile \cite{Fabijanska:2021:GNN} and  will certainly have a big impact on systems biology research. A major difficulty with these approaches is that they are considered black-box models. The decisions they make cannot be traced back precisely, therefore medical doctors may not trust decisions made by black-box models. To address this shortcoming, first explainable GNN methods are available and are currently under strong development. Examples include GNNexplainer \cite{ying2019gnnexplainer}, PGExplainer \cite{luo2020parameterized}, and GNN-LRP \cite{schnake2020higher}. GNNExplainer provides \textit{local} explanations for predictions of any graph-based model. It can be applied to node classification as well as graph classification tasks. PGExplainer is a parameterized modification of the GNNexplainer. In contrast to the GNNexplainer, it provides explanations on a model level; especially useful for graph classification tasks. The GNN-LRP approach is derived from higher-order Taylor expansions based on Layer-wise relevance (LRP). It explains the prediction by extracting paths from the input to the output of the GNN model that contributes the most to the prediction. These paths correspond to \textit{walks} on the input graph. GNN-LRP was developed for explanations on the node level but was recently modified to also work for graph classification in a special application set-up \cite{chereda2021explaining}.  

The aforementioned methods may facilitate the discovery of disease-causing regions within biological networks, but not directly can be used for network module \textit{selection} purposes. Explainers on GNNs usually highlight the importance of edges, nodes or walks within the network, but they do not rank whole subnetworks by their importance. Also, methods for explanations are often black-box models themselves, and thus may not be trustworthy either. In addition, standard GNN architectures cannot handle multi-graphs which may hinder the analysis for incomplete data, which is often the case in the biomedical domain.

At the same time, only a few FS methods are applicable on network structured input data. Some developments in this direction are realized within the R-packages glmnpath \cite{chen2015glmgraph} %and know-GRRF \cite{guan2020dynamic}. 
The implemented method uses a regularization term to incorporate network topology into the feature selection process. The underlying algorithm, however, does not function on the network itself.

In this work, we present a greedy decision forest for the selection of disease network modules.
The proposed algorithm directly functions on a network by sampling node features by random walks. A set of decision trees are derived from the visiting nodes, ultimately forming the decision forest. A greedy process on the selection process of the decision trees is initiated, and the algorithm converges to a set of subnetworks. In the following we will use the term "network module" and "subnetwork" interchangeable. 

Our proposed methodology %is a "glass-box" approach \cite{HolzingerEtAl:2017:glassbox} and 
naturally handles multi-modal data with a high degree of interpretability, thereby perfectly suited for multi-omics types of analyses in the biomedical domain. %In principle, the addressed problem could also be solved by using GNNs and explainable GNN for subnetwork detection. However, results may not be as interpretable as the glass-box approach we provide. Moreover, it encourages both the domain expert and the AI system's designer to pursue actions for the improvement of performance and explainability.  

%TODO: Is it an a-posteriori? Is it suitable for 

%%%%%%%%%%%%%%%%%%
\section{Greedy Decision Forest}

% Figure (Graphical abstract)
\begin{figure*}[h!]
  %\centering
  \includegraphics[scale=0.50]{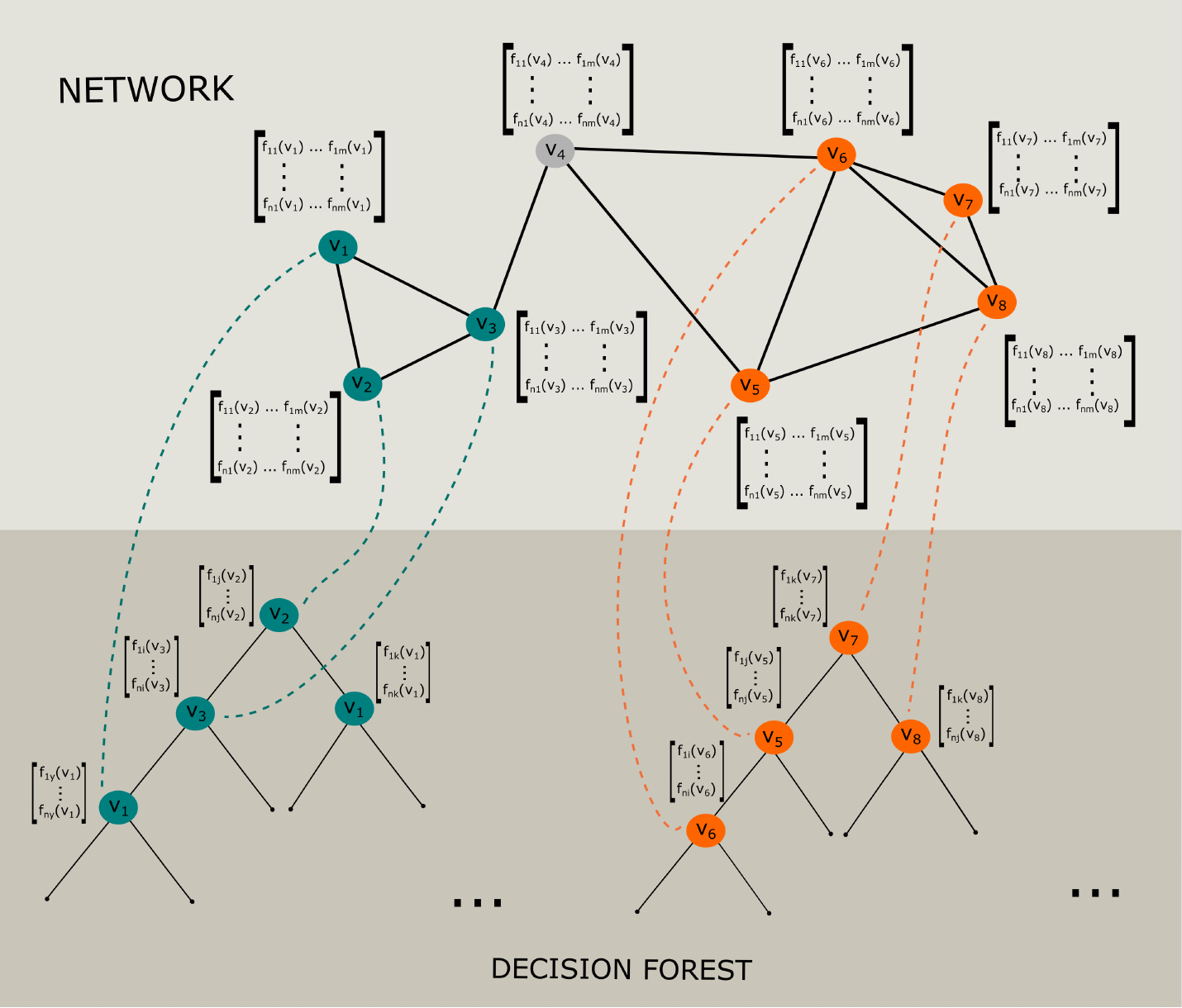}
  \caption{Deriving a Decision Forest from a network comprising multi-modal node feature vectors. Displayed is the network at the top and the derived decision forest at the bottom. Each node of the network comprises $m$ node feature vectors of length $n$. Each random walk creates a $mtry$ set of feature vectors used for building a decision tree.}
\end{figure*}

\noindent The core functionality of our proposed network module detection approach includes a tree-building process derived from a network $G=(V,E)$. Each node $\mathbf{v}\in V$ comprises node feature vectors $f(\mathbf{v})$ of arbitrary dimensions (see Fig. 1). In order to build a single tree we randomly select a node $\mathbf{v} \in V$ located on the network $G$. From that node a random walk is initialized. The depth of this random walk is set to $\sqrt{\#V}$, which is the squared root of the total number nodes. This value is frequently used as a rule-of-thumb, and a default value in popular R-packages like ranger \cite{wright2015ranger} and randomForest \cite{liaw2002classification}. It is the so-called $mtry$ parameter, and it specifies the number of features a decision tree includes. In our case, features of visited nodes by the random walk are in the $mtry$ set.    
The described procedure is repeated until \textit{ntree} decision trees are generated. Figure 1 illustrates this process. A network is shown with eight nodes. Each of these nodes has $m$ node feature vectors of length $n$, where $n$ is the number of samples and $m$ is the number of modalities. Two random walks of depth four are highlighted. Each of these walks forms a Decision Tree (blue and orange nodes). It should be noted, that the random walk may visit the exact same node multiple times. So is the case for the first random walk. The derived Decision Tree contains the node $v_{1}$ twice. Thus, the associated features are used twice for splitting a tree node (Fig.1: blue Decision Tree). Node $v_{4}$ was not captured by the random walks.

Based on the network-derived decision trees we start the proposed greedy algorithm for network module selection. At each greedy step, $ntree$ decision trees are generated, while the selection of network nodes and their corresponding features depend on the outcome of previous iterations. High performing trees remain in the \textit{ntree} set while low performing trees and their corresponding nodes and node features are eliminated.

% You could \cite{Kochenderfer:2019:AlgorithmsOptimization} https://en.wikipedia.org/wiki/Fitness_proportionate_selection, https://en.wikipedia.org/wiki/Tournament_selection 

In the following, we will introduce necessary notations and describe the greedy algorithm in great detail.

% Subsection
\subsection{Network module selection using a greedy decision forest}

\noindent We define a Decision Tree classifier as $T(x;\Theta, X)$, where $\Theta$ consists of a set of split rules and $X$ includes the variables/features used for splitting. Given an input vector $x$, $T(x;\Theta, X)$ assigns a given data point to a specific class. An ensemble of Decision Tree classifiers is called Decision Forest (DF). It is defined as $\{T_{k}(x,\Theta_{k}, X_{k}), k=1,\ldots, ntree\}$, where $X_{k}$ is a set of randomly selected features from the input feature space the $k$-th Decision Tree $T_{k}$ is based on. 

Let us further assume a graph $G=(V,E)$ is specified. The nodes $v\in V$ represent the features in $X$ and the edges $(v_{i},v_{j}) \in E$ are reflecting any kind of dependency of these features. This graph for example could be a knowledge graph, which connects node features with domain-specific and domain-relevant relationships. An example of such graph is a Protein-Protein network, where lab-validated functionally interacting genes are connected.

Here, we propose to restrict the samples $X_{k}$ to be neighboring nodes within graph $G$. This regularization ensures that related features are located on the same Decision Tree. As a consequence, classifications made by this tree model might be more reliable and interpretable for the domain expert. Accordingly, our proposed Decision Forest is defined as $\{T_{k}(x,\Theta_{k}, X^{G}_{k}), k=1,\ldots, ntree\}$, where $X^{G}_{k}$ is a set of features determined by a random walk on graph $G$. Starting from that Decision Forest we execute the proposed greedy steps for tree-based module selection (see Algorithm 1).

%%%% Algorithm 1 
\begin{algorithm}[H]
\SetAlgoLined
%\KwResult{Write here the result }
 Given a Decision Forest: $\{T_{k}(x,\Theta_{k}, X^{G}_{k})\}$\;
 Given a graph: $G=(V,E)$\;
 $k=\{1,\ldots, ntree\}, t=1$\;
 $mtry_{k}=\sqrt{\#V}$\;
 $X^{G}_{k}[t=1]=X^{G}_{k}$\;
 
 \While{$t \leq niter$}{
 
 \For{$k\gets1$ \KwTo $ntree$ }{
  $Perf(T_{k}[t])= Performance~of~T_{k}(x;\Theta_{k}, X^{G}_{k})$\;

  \eIf{$Perf(T_{k}[t]) \leq Perf(T_{k}[t-1])$}{
   $T_{k}[t]=T_{k}[t-1]$\;
   $X^{G}_{k}[t]=X^{G}_{k}[t-1]$\;
   }{
    $mtry_{k}[t]--$\;
    $X^{G}_{k}[t]=RandomWalk(G | X_{k}, mtry_{k}[t]) $\;
   }
   }
  $Sample~ntree~trees~according~to~Perf\{T_{k}[t]\}$\;
  $t++$\;
 }
 \caption{Greedy Decision Forest}
\end{algorithm}

\noindent For each greedy step $t$ we calculate the performance for all $k\in \{1,\ldots,ntree\}$ Decision Trees based on the \textit{out-of-bag} samples (Algorithm 1: line 8). Here, we use ROC-AUC as a performance estimate, but this could be replaced by any type of performance measure such as accuracy (in the case of a balanced dataset) or mutual information \cite{MacKay:2003:InformationTheory}. In case the \textit{out-of-bag} performance of the $k$-th Decision Tree $Perf(T_{k}[t])$ at greedy step $t$ is lower than the performance of the Decision Tree from the previous iteration ($t-1$), the suggested Decision Tree and the corresponding node features are dropped. In case the Decision Tree provides better performance, a random walk on a subgraph, specified by the features in $X_{k}[t]$ at greedy step $t$, is initialized (Algorithm 1: line 14). Note, the depth of this walk is now decreased by one (Algorithm 1: line 13). Consequently, the algorithm aims for a minimal set of tree features while not decreasing its performance. After updating the Decision Trees, we sample a new set of $ntree$ Decision Trees according to their \textit{out-of-bag} performance values which initiates the mentioned selective process (Algorithm 1: line 17). The algorithm repeats the aforementioned procedure and terminates after $niter$ greedy steps. The selected modules are represented by the node features $X^{G}_{k}$ at iteration $t=niter$. 

The feature space $X$ can naturally be extended to a multi-modal representation. It may simply consist of a set of matrices instead of a single matrix, representing the samples as rows and the features as columns. Figure 1 illustrates this scenario. Each node is represented by a feature matrix. Each column of this matrix is a feature vector from a specific modality, all associated with the same node. The tree building process is free to pick features from any of these given modalities for splitting a tree node, as long it is associated with the same network node. 

Decision Forests are particularly suited for the multi-modal case, because DFs are scale invariant. There is no need for data normalization prior to execution and thus DFs provide an ideal framework for the analysis of heterogeneous input data. At the same time the node-split rules of a DF can be read out and therefore it is straightforward to re-trace the contribution of each modality to the final outcome. 

%TODO: Create an example to show it theoretically.

%%%% Module Importance 
\subsection{Network module importance scores}

\noindent We compute an importance score for each of the selected modules. The proposed estimate is based on the network module \textit{out-of-bag} performance $Perf(T_{k})$ at the $niter$ greedy step, which is the performance of the Decision Tree associated with that specific module. However, $Perf(T_{k})$ estimate of a module does not depend on the number of features used by the derived Decision Tree. Thereby, redundant features do not affect the performance in a negative way. Here, we aim to infer the smallest possible module with maximal performance. Thus, an additional importance measure is needed to account for the importance of the actual edges forming the module. Edge importance is calculated as

\begin{equation}
    IMP_{e}((\mathbf{v}_{i},\mathbf{v}_{j}) \in E)):= \sum_{t}^{niter}\sum_{k}^{ntree} Perf(T_{k}(\{\mathbf{v}_{i},\mathbf{v}_{j}\}\in X_{k}^G)[t]).
\end{equation}

\noindent The above equation assigns the performance of a module $X_{k}^G$ to the edges $(\mathbf{v}_{i},\mathbf{v}_{j})$ forming it. This is accomplished for all trees generated during the greedy steps $t$. Consequently, we not only account for the performances but also for the number of times an edge is part of the $ntree$ set. As a result, less important edges are purified during the selective greedy process.

\noindent The normalized edge importance of a module is calculated as
\begin{equation}
    \overline{IMP_{e}}(X_{m}^G):= \frac{\sum_{i,j} IMP_{e}((\mathbf{v}_{i},\mathbf{v}_{j})\in X_{m}^G)}{\# (\mathbf{v}_{i},\mathbf{v}_{j})\in X_{m}^G},
\end{equation}

\noindent where $m\in \{1,\ldots,ntree\}$ and $(\mathbf{v}_{i},\mathbf{v}_{j}) \in E$, where $i,j \in \{1, \ldots, \#V\}$. The nominator reflects the cumulative edge importance score within the $m$-$th$ module $X_{m}^G$, and is normalized by the number of edges forming that module.

\noindent Finally, module importance is calculated as
\begin{equation}
    IMP_{m}(X_{m}^G):= \overline{IMP_{e}}(X_{m}^G) + Perf(T_{m}),
\end{equation}

\noindent where $Perf(T_{m})$ is the \textit{out-of-bag} performance of the $m$-th DT associated with the module $X_{m}^G$. We utilize the $IMP_{m}$ to rank the obtained modules $X_{m}^G$ after $niter$ greedy steps.

% Subsection
\subsection{Node feature importance scores}
Our proposed Greedy Decision Forest allows to retrace from which modality the features were sampled to form the selected modules. The importance of these features can be calculated by standard tree-based importance measures, such as as the Gini impurity index \cite{Witten:2005:DataMining}. The Gini index at tree node $\mathbf{v}^{t}$ is

\begin{equation}
    Gini(\mathbf{v}^{t})=\sum_{c=1}^{C}\hat{p}_{c}^{\mathbf{v}^{t}}(1-\hat{p}_{c}^{\mathbf{v}^{t}}),
\end{equation}

where $\hat{p}_{c}^{\mathbf{v}^{t}}$ is the proportion of samples belonging to class $c$ at tree node $\mathbf{v}^{t}$. The Gini information gain obtained by a feature $X_{i}$ for splitting node $\mathbf{v}^{t}$ is the difference between the Gini impurity at node $\mathbf{v}^{t}$ and the weighted average of impurities at each
child node of $\mathbf{v}^{t}$. The Gini information gain is defined as

\begin{equation}
    Gain(X_{i},\mathbf{v}^{t})=Gini(\mathbf{v}^{t})-w_{L}Gini(\mathbf{v}^{t}_{L},X_{i})-w_{R}Gini(\mathbf{v}^{t}_{R},X_{i}),
\end{equation}

\noindent where $\mathbf{v}^{t}_{L}$ and $\mathbf{v}^{t}_{R}$ are the left and right child nodes of $\mathbf{v}^{t}$, respectively, and $w_{L}$ and $w_{R}$ are the proportions of $c$-class instances assigned to the left and right child nodes. At each node $\mathbf{v}^{t}$, a set of features (the $mtry$ set), and the feature with the maximum $Gain(X_{i}, \mathbf{v}^{t})$ is used for splitting the node. 
We calculate the importance of a feature within a detected module $X_{m}^G$ as 

\begin{equation}
IMP_{f}(f(\mathbf{v}_{i}\in V))
=\sum_{t}^{niter}\sum_{k}^{ntree} Gain(f(\mathbf{v}_{i})\in X_{k}^G)[t],
\end{equation}

\noindent and it is normalized as

\begin{equation}
\overline{IMP_{f}}(f(\mathbf{v}_{i})\in X_{m}^G)
= \frac{IMP_{f}(f(\mathbf{v}_{i})\in X_{m}^G)}{niter\cdot ntree},
\end{equation}

\noindent where $f(\mathbf{v}_{i})$ refers to a feature vector associated with a network node $\mathbf{v}_{i}$. Similar to the calculation of the edge importance $IMP_{e}$, we measure the importance of a feature by accounting for the number of times it is a member of the $ntree$ set of modules during the greedy process. 

%\begin{equation}
%    Imp_{i} = \frac{1}{ntree}\sum_{v\in S_{X_{i}}} Gain(X_{i},\mathbf{v}),
%\end{equation}
%where $S_{X_{i}}$ is the set of nodes split by $X_{i}$ in the RF comprising $ntree$ trees.

%TODO: Please provide results for experiments \textbf{without} the $Perf(T_{m})$ term. 

%Subsection
\subsection{The greedy decision forest as a predictive classifier}

\noindent We also consider our proposed GDF to be a machine learning model calculating predictions based on multi-modal data. We can explain the local behaviour of this model using SHapley Additive exPlanations algorithm (SHAP)~\cite{lundberg2017unified}, which is a state-of-the-art approach for explaining any predictive classifier~\cite{JMLR:v11:strumbelj10a, vstrumbelj2014explaining}. A tree-based structure of GDF allows obtaining informative local and global explanations using the efficient TreeSHAP algorithm~\cite{lundberg2020local}. Moreover, we aim to introduce a corresponding node feature importance score based on SHAP. This alternative further extends our understanding of GDF and promotes it as a glass-box approach for biomedical knowledge discovery and decision making.
 
SHAP values $sv_j(x)$ for a given decision forest $DF$, an observation $x$ and feature $j$ are defined as

\begin{equation}
sv_j(x)  = 
\frac 1{|V|} \sum_{s=0}^{|V|-1} \sum_{
\substack{
S \subseteq V\setminus \{j\} \\ |S|=s
}}  {{|V|-1}\choose{s}}^{-1} \left( e ^{S \cup \{j\} }(x) - e ^{S}(x)\right),
\end{equation}

where  $e^{S}(x)$ stands for an expected value for model prediction for variables with set variables with indexes from set $S$, i.e. $e^{S}(x) =  E [DT(x^*)|\forall_{i \in S} x_i = x_i^*]$. Intuitively, this is the average change in the model prediction if we set a value in variable $j$.

Shapley values are additive, i.e.

\begin{equation}
DF(x) = sv_0 + \sum_{j} sv_j(x), % very simply speaking, we could give an extended formula in the form of sv_j = ...
\end{equation}
where $sv_0$ is the average baseline probability and $sv_j(x)$ equals the $j$-th feature's attribution to the prediction $DF(x)$. Consecutively, SHAP importance for feature $j$ can be defined as an average of the absolute SHAP values per feature across the observations in data
\begin{equation}
SVIMP_j = \frac{1}{|X^G_m|}\sum_{x \in X^G_m} |sv_j(x)|.
\end{equation}
We note that this is a widely-used heuristic approach, while other aggregations might also be used to quantify feature importance. The main property of SHAP is their additivity with respect to features, which we utilize to aggregate the attributions across multi-modal data. Using the SHAP framework, node feature importance is calculated as
\begin{equation}
SVIMP_f = \sum_{j \in f(v_i)} SVIMP_j.
\end{equation}

%\begin{figure*}[h!]
% \centering
%  \includegraphics[width=1\textwidth]{Main/shap_demo.png}
%  \caption{~Exemplary explanations.}
%\end{figure*}

%%%%%%%%%%%%%%%%%%%%%%%
\section{Synthetic data sets}
We have exercised our approach on synthetic Barabasi networks \cite{barabasi2004network}. Barabasi networks were developed to reflect real-world biological networks, where subsets of nodes (e.g genes) are strongly connected and organized as communities. They are scale-free networks and are generated using a preferential attachment mechanism. 
For each simulated network we randomly selected four connected nodes $\mathbf{v}_{1}, \mathbf{v}_{2}, \mathbf{v}_{3},$ and $\mathbf{v}_{4}$, which combined with their associated edges form a functional module.
Each node is linked to a feature vector comprising binary feature values of 1000 samples. The feature values of all nodes are uniformly distributed, while the selected nodes have the following functional relationship

\begin{equation}
    Module(V,E):= [f(\mathbf{v}_{1}) \wedge f(\mathbf{v}_{2})] \oplus [f(\mathbf{v}_{3}) \wedge f(\mathbf{v}_{4})],
\end{equation}
where $f(\mathbf{v})$ is the feature vector associated with node $\mathbf{v}$.

\noindent We have generated a Barabasi network comprising $\#V=30$ nodes with a power of the preferential attachment of $1.2$ using the R-package igraph \cite{igraphPackage}. The selected nodes of the functional module are $\mathbf{v}_{12}, \mathbf{v}_{15}, \mathbf{v}_{16},$ and $\mathbf{v}_{30}$ and form a sub-graph. Figure 1 shows the resulting Barabasi Network and the top-1 module correctly verified by our approach (highlighted in red). The thickness of the edges reflects the importance of the edges determined by $IMP_{e}$.

% Figure
\begin{figure*}[h!]
 \centering
  \includegraphics[scale=0.70]{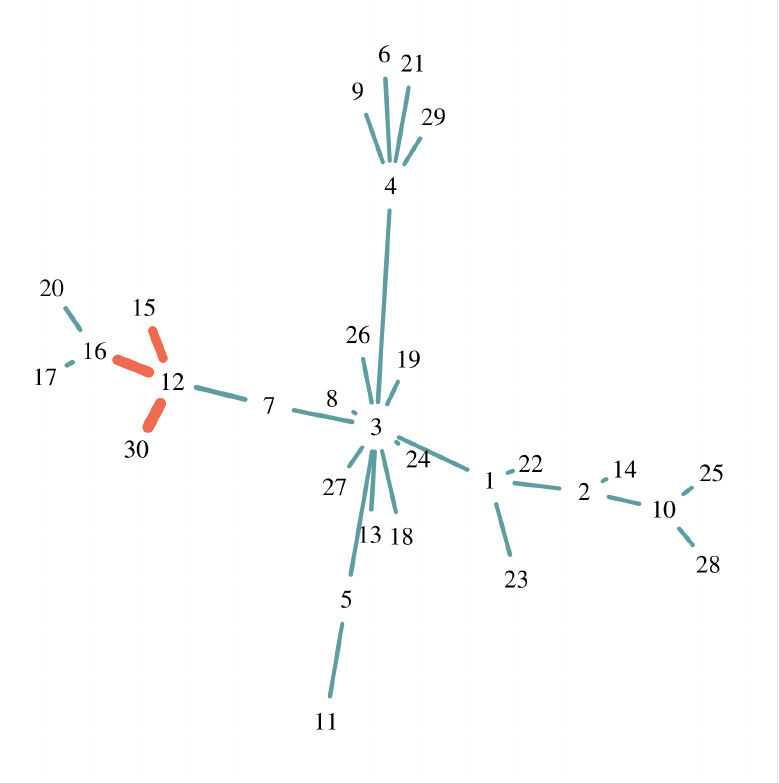}
  \caption{Simulated Barabasi network. The correctly identfied subnetwork (\{$\mathbf{v}_{12}, \mathbf{v}_{15}, \mathbf{v}_{16},$, $\mathbf{v}_{30}$ \}) is highlighted in red. The thickness of the edges is reflected by the  $IMP_{e}$ edge importance values. The \textit{out-of-bag} performance of the detected module is $Perf(T_{m})=1$, and the normalized edge importance score is  $\overline{IMP_{e}}=0.67$. The overall importance score thus is $IMP_{m}=1.67$.}
\end{figure*}

Furthermore, Table 1 lists the selected modules sorted by their module importance scores ($IMP_{m}$). Overall, six unique modules were detected. As can be seen from Table 1, the first three modules have the same \textit{out-of-bag} module performance ($Perf(T_{m})=1$). Their associated DTs perform equally well with an accuracy of $AUC=1$. However, these modules differ with regard to their average edge importance ($\overline{IMP_{e}}$). Redundant node features 17 (in case of module 2) and feature 20 (in case of module 3) decrease the average edge importance scores. If we would judge the detected modules based on their module performance $Perf(T_{m})=1$ only, it may result in top-ranked modules including redundant features. On the other hand, if we would exclusively rely on the edge importance $\overline{IMP_{e}}$, we may miss important features. In our example this would make module $\{12, 16, 30\}$ ranked second (see Table 1). In that case, we miss relevant node features associated with node $v_{15}$. The combination of both, $\overline{IMP_{e}}$ and $Perf(T_{m})$ provides a minimal set of most important node features.

\begin{table}[ht]
\centering
\caption{Detected modules from the Barabasi network shown in Figure 2.}
\begin{tabular}{ c  l  c  c  c}
  \hline
  RANK & MODULES & $\overline{IMP_{e}}$ & $Perf(T_{m})$ & $IMP_{m}$ \\
  \hline			
  1 & \{12, 15, 16, 30\}        & 0.67 & 1 & 1.67 \\
  2 & \{12, 15, 16, 17, 30\}    & 0.57 & 1 & 1.57 \\
  3 & \{12, 15, 16, 20, 30\}    & 0.56 & 1 & 1.56 \\
  4 & \{12, 16, 30\}            & 0.64 & 0.81 & 1.45 \\
  5 & \{12, 15, 30\}            & 0.59 & 0.80 & 1.39 \\
  6 & \{12, 16, 17, 30\}        & 0.51 & 0.80 & 1.31 \\
  \hline  
\end{tabular}
\end{table}

% Table
%       Module #EDGE_IMP       AUC
%1    12 15 16 30 0.6736688 1.0000000
%2       12 16 30 0.6383748 0.8066523
%3       12 15 30 0.5931836 0.7977642
%4 12 15 16 17 30 0.5662249 1.0000000
%5 12 15 16 20 30 0.5595749 1.0000000
%6    12 16 17 30 0.5128934 0.8018943

In a further analysis we studied the effect of the parameter $niter$ on the performance of our module selection method. We have generated a Barabasi network comprising 50 nodes and executed the module selection procedure 100 times. For each run the topology of the barabasi network is the same, but we varied the selected nodes as well as the binary feature vectors. We report on the top-1 coverage, which in our case is the number of times the module of interest is ranked first. In addition, we report on the number of unique modules out of the $n.tree$ set after $niter$ greedy steps, and the overall \textit{out-of-bag} performance of the greedy Decision Forest classifier. Results are shown in Fig. 3. 

\begin{figure*}[h!]
  \centering
  \includegraphics[scale=0.70]{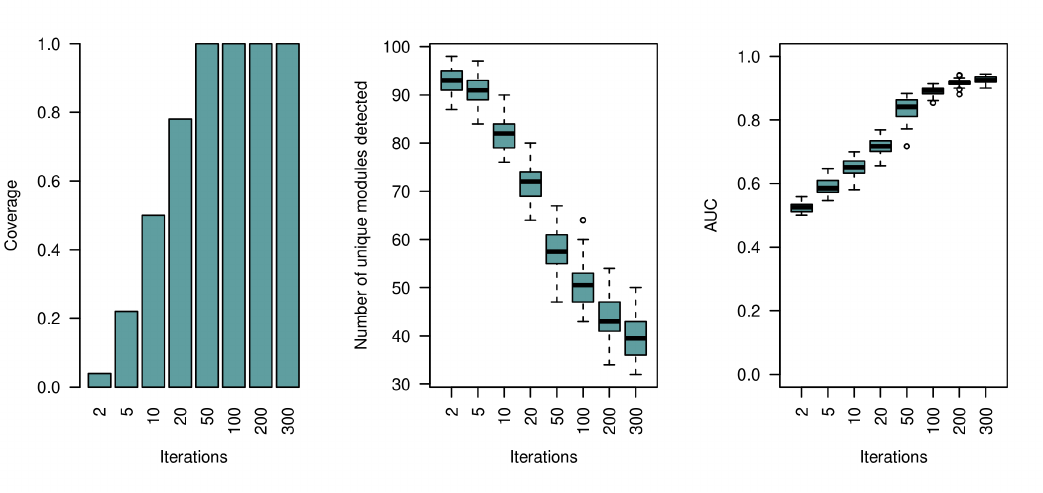}
  \caption{\textit{Single-Modal} simulation results on Barabasi networks. We varied the number of greedy iterations ($n.iter$) and calculated the number of times the selected module is ranked first according to our proposed module importance score $IMP_{m}$ (left panel). Displayed are the number of unique modules within the $n.tree$ module set after termination of the greedy process (middle panel). The \textit{out-of-bag} performance of the greedy Decision Forest classifier is shown in the right panel. For each run the topology of the Barabasi network, the feature values, and the selected subnetwork is the same. The simulated Barabasi networks comprise 50 nodes.}
\end{figure*}

\noindent The target network module is consistently ranked first ($coverage=1$) after 50 greedy iterations. Also, the performance of the Decision Forest classifier stabilizes at $AUC~=0.9$ after 50 greedy runs. The number of unique modules decreases with the number of iterations and starts to converge at 200 iterations. We observed that the lower-ranked modules are forming graphs which all include the module ranked first as a sub-graph (similar to what can be seen from Table 1).      

\noindent Each node may include heterogeneous multi-modal features. This is typically the case in integrative multi-omic studies. Possible features may include gene expression levels, micro-RNA, and DNA Methylation data for the same set of patients. In order to test the general applicability of our approach for these type of applications we define a multi-modal XOR module on a Barabasi network as follows:

\begin{equation}
    Module(V,E):= [f_{a}(\mathbf{v}_{1}) \wedge f_{b}(\mathbf{v}_{2})] \oplus [f_{a}(\mathbf{v}_{3}) \wedge f_{b}(\mathbf{v}_{4})],
\end{equation}
where $f(\mathbf{v})$ is the feature vector associated with node $\mathbf{v}$, and $a$ refers to the features of the first modality and $b$ to the second.

\begin{figure*}[h!]
  \centering
  \includegraphics[scale=0.70]{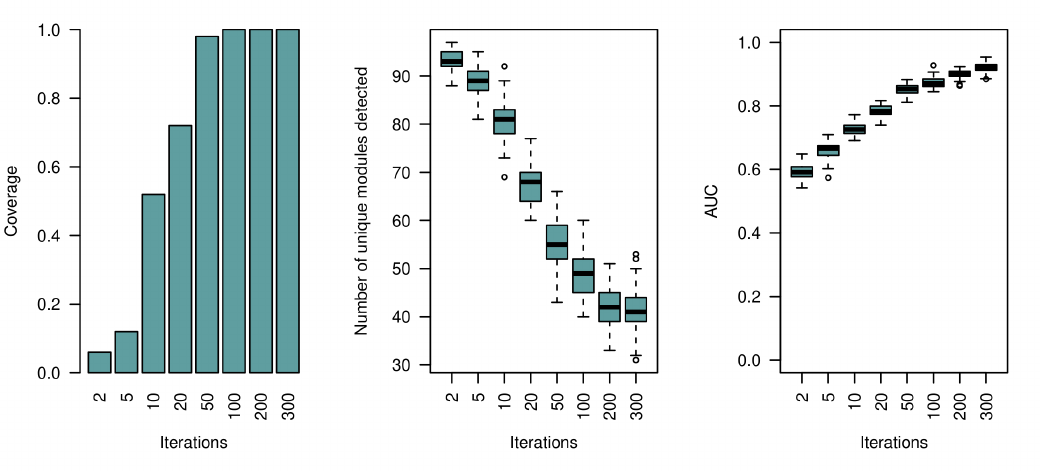}
  \caption{\textit{Multi-Modal} simulation results on Barabasi networks. We varied the number of greedy iterations ($n.iter$) and calculated the number of times the selected module is ranked first according to our proposed module importance score $IMP_{m}$ (left panel). Displayed are the number of unique modules within the $n.tree$ module set after termination of the greedy process (middle panel). The \textit{out-of-bag} performance of the greedy Decision Forest classifier is shown in the right panel. For each run the topology of the Barabasi network, the feature values, and the selected subnetwork is the same. The simulated Barabasi networks comprise of 50 nodes.}
\end{figure*}

\begin{figure*}[h!]
  \centering
  \includegraphics[scale=0.70]{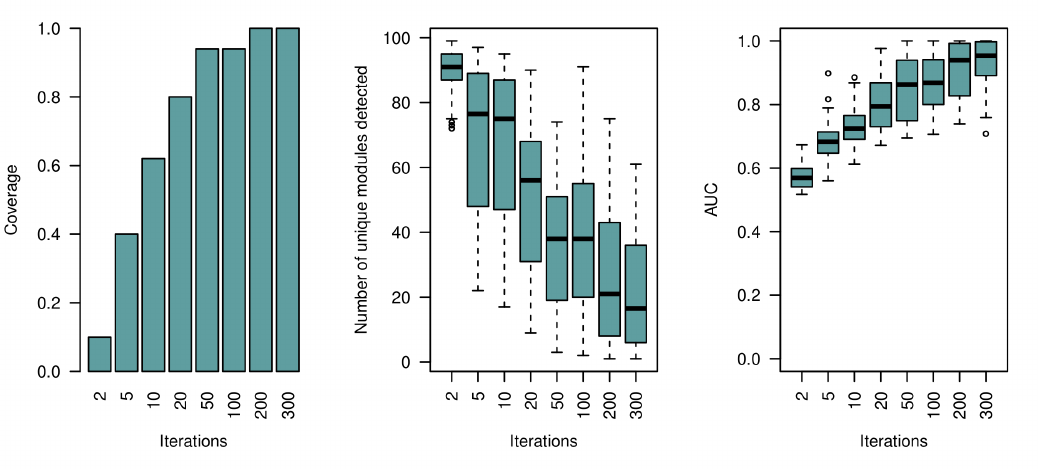}
  \caption{\textit{Multi-Modal} simulation results on \textit{variable} Barabasi networks. We varied the number of greedy iterations ($n.iter$) and calculated the number of times the selected module is ranked first according to our proposed module importance score $IMP_{m}$ (left panel). Displayed are the number of unique modules within the $n.tree$ module set after termination of the greedy process (middle panel). The \textit{out-of-bag} performance of the greedy Decision Forest classifier is shown in the right panel. For each run the topology of the Barabasi network, the feature values, and the selected subnetwork \textit{varies}. The simulated Barabasi networks comprise of 50 nodes.}
\end{figure*}

\noindent Fig. 4 shows the results when the Barabasi network structure is the same in all iterations. The results are very similar to what we have seen in the single-modal case (Fig. 3). This observation indicates that the proposed algorithm is fully capable of detecting important sub-networks even when the relevant information is distributed across multiple modalities.

In a additional investigation, we varied the topology of the Barabasi networks, as well as their corresponding binary node features (Fig. 5). Results confirm that the proposed Greedy Decision Forest can be efficiently applied to multi-modal feature inputs. The coverage is comparably high and is 1 starting with 200 greedy iterations (see Fig. 5). However, both, the \textit{out-of-bag} $AUC$ values and the number of detected unique modules have a wider distribution (Fig. 5). Thus, ee can conclude that the performance of our proposed algorithms depends on the topology of the barabasi networks (see Fig.4 versus Fig. 5).

%%%%%%%%%%%%%%%%%%%%%%%%%%%%%%%%%%%%%%%
\section{Application to multi-omics TCGA cancer data}

We showcase the applicability of our approach on a Protein-Protein Interaction Network (PPI) for the detection of disease modules. The PPI network was retrieved from the STRING database \cite{mering2003string}. We only kept high confidence interactions (a \textit{combined score} within the upper 0.95 quantile). We filtered for relevant cancer genes as specified in \cite{schulte2021integration}. This resulted in 13,218 relevant genes and 6,926,452 edges. We enriched each node of the PPI network by multi-omic features from kidney survival and non-survival patients. The omic features were extracted from \textit{http://linkedomics.org/} \cite{vasaikar2018linkedomics}. We selected gene expression (mRNA) and DNA Methylation data for the same set of patients. 

Two experiments were conducted. First, we investigated the capability of the proposed approach to predict the survival status of patients suffering kidney cancer, to infer potential subnetworks for risk assessments.

In a second experiment, our aim was to detect kidney-specific disease modules. For this analyses we randomly selected patients suffering lung, ovarian, and breast cancer and assigned these patients to a non-kidney cancer group. Here the goal was to detect modules as potential biomarkers specific to kidney cancer. After harmonization with the reduced PPI network, we ended up with 6249 network nodes (genes) for the survival analyses, and 3374 genes for the cancer type experiment (see Table 2).

We applied a 80\%-20\% train-test split and initiated 1000 random walks (n.tree=100) on the PPI network based on the train set. The number training and test samples are shown in Table 2.

\begin{table}[ht]
\centering
\caption{Experimental set-up (\textit{s=survived, ns=non-survived, rand=random cancer type)}}
\begin{tabular}{l c  l  l }
  \hline
   Experiment & Genes & Patients (train set) & Patients (test set) \\
  \hline			
  Survival status   & 6249  &  85 (s), 79 (ns)  & 18 (s), 24 (ns)  \\
  Cancer type       & 3374  & 162 (kidney), 158 (rand) & 38 (kidney), 42 (rand)   \\
  \hline  
\end{tabular}
\end{table}

The depth of the random walks, capturing the nodes for tree building, was set to an initial value of $mtry=30$. Thus, we were interested in disease modules comprising less than 30 genes. The derived 1000 Decision Trees we then let evolve $n.iter=100$ greedy iterations. The greedy process terminated after 51000 trees. The predictive performance of our greedy DF classifier can be obtained from Table 3. It is based on the predictions of the last 1000 trees (\textit{n.tree}) of the decision forest, which include the selected modules.

\begin{table}[ht]
\centering
\caption{Performance of the greedy decision forest}
\begin{tabular}{l c  l  c  c  c}
  \hline
  Experiment & Sensitivity & Specificity & Recall  & Precision & Accuracy \\
  \hline			
  Survival status   & 0.75 &  0.78 & 0.75 & 0.82 & 0.76 \\
  Cancer type & 0.86 &  0.74 & 0.86 & 0.78 & 0.80 \\
  \hline  
\end{tabular}
\end{table}

We compared the predictive performance of our greedy DF classifier with a standard Random Forest and a fully connected Neural Network. For these classifier we do not incorporate any domain-knowledge about the interaction and relatedness of genes. We have repeated the 80\%-20\% train-test split 20 times and report on the min, median, and max accuracy (see Table \ref{tab:Comparison}).

% Table (Apllication Results)
%Table 2
\begin{table}[h]%[!b]%[h] %[t]%[h!]
  %\begin{center}
    \caption{Classification accuracy (min/median/max) on the Cancer Type experiment.}
    %\label{tab:table1}
    % \centering
    %\begin{tabular}{p{0.10\linewidth}p{0.10\linewidth}p{0.08\linewidth}p{0.08\linewidth}p{0.08\linewidth}p{0.08\linewidth}p{0.08\linewidth}p{0.08\linewidth}p{0.08\linewidth}}
    \begin{tabular}{l l c c c} % <-- Alignments: 1st column left, 2nd middle and 3rd right, with vertical lines in between
      \textbf{Classifier} & \textbf{Cancer} & \textbf{mRNA} & \textbf{DNA Methylation} & \textbf{Multi-omics} \\
      \hline
      DFNET (ours) & KIRC & 75/78/84 & 78/84/86 & 79/85/89 \\
      & BRCA & 62/70/78 & 69/70/83 & 73/76/83  \\
      & LUAD & 80/87/96 & 79/83/88 & 80/86/91 \\ %\hline %\hline
     NN & KIRC & 82/86/91 & 82/87/92 & 82/86/93 \\
       & BRCA & 65/74/85 & 71/77/82 & 56/78/86  \\
     & LUAD & 81/87/91 & 83/87/92 & 82/88/95 \\ %\hline
   %  DT & KIRC & 74/85/91 & 77/87/94 & 80/87/92 \\
%      & BRCA & 61/69/78 & 66/76/85 & 67/77/85  \\
 %    & LUAD & 71/80/89 & 83/88/93 & 78/88/95 \\ %\hline
     RF    & KIRC & 80/83/90 & 82/87/94 & 81/88/92  \\
       & BRCA & 65/76/86 & 67/79/85 & 71/78/82 \\
      & LUAD & 77/87/92 & 82/89/94 & 85/88/94    \\ \hline
      
    \end{tabular}
%\begin{tablenotes}
%\item [ ] \footnotesize{--}
%\end{tablenotes}
  %\end{center}
   \label{tab:Comparison}
\end{table}

We learned that the alternative approaches performed equally good as the proposed greedy decision forest, in terms of model performance. This is an intriguing result since we aim for increased interpretability while maintaining predictive performance. 
The main focus of this paper, however, is module selection, and classical machine learning approaches do not provide such selective procedure. %We will report on an comparison with RF regarding feature selection in the next paragraph. 
Our DF classifier comprises of decision trees with functional related biological entities and thus can be better interpreted by the domain expert. %Nevertheless, we plan an in-depth analyses of the PPI Network. Biases in the construction of these networks may lead to spurious results. We plan further investigations in this direction for advanced applications of our proposed methodology.

Ultimately, for the survival analysis, the detected modules/trees include about 10-20 genes. The top-ranked module, according to $IMP_{m}$, comprises 15 genes (see Fig. 6). The module has an \textit{out-of-bag} accuracy of $0.72$. The test set performance of that module can be seen in Table 5. 

\begin{figure*}[h!]
  \centering
  \includegraphics[scale=0.44]{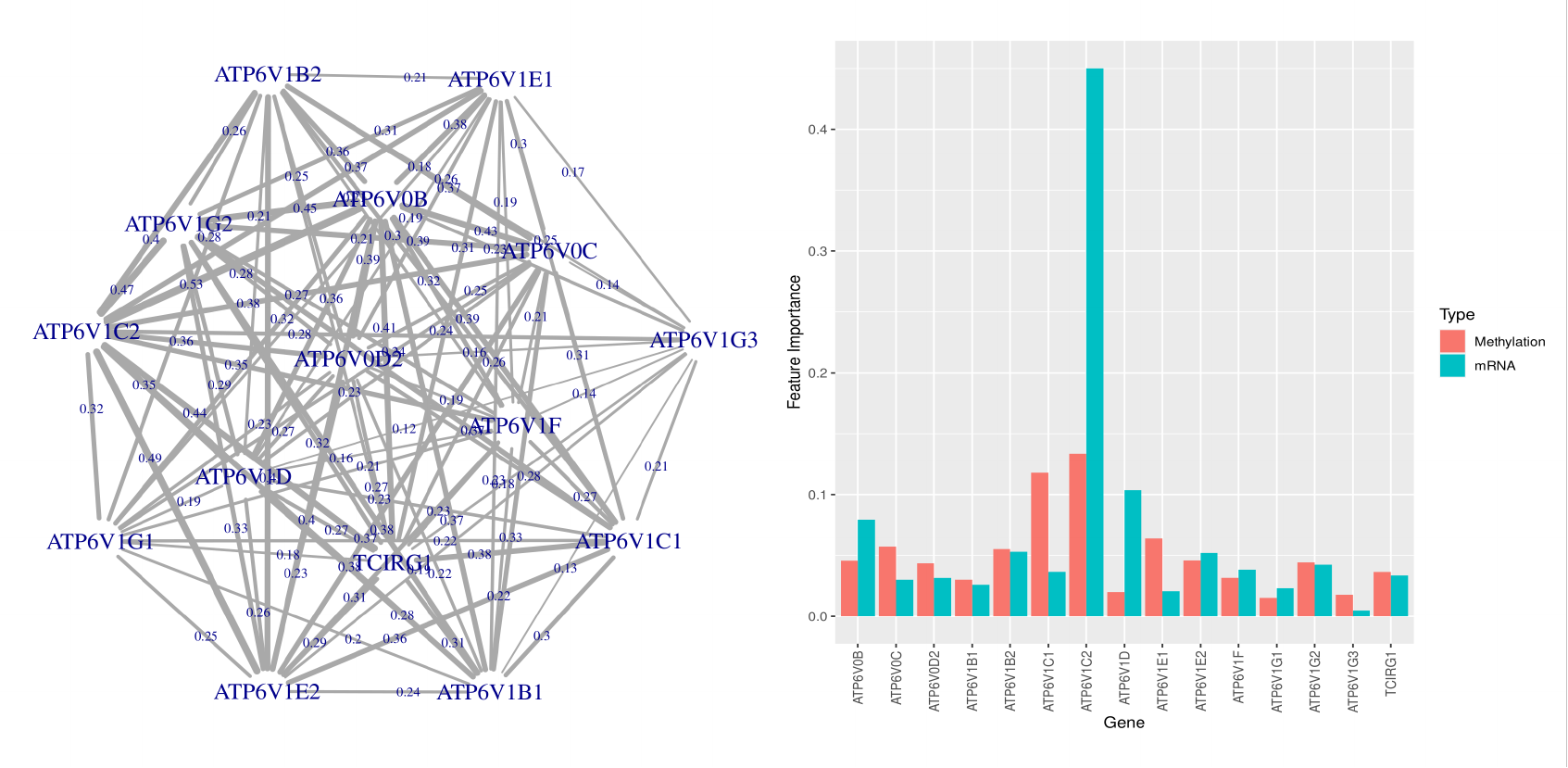}
  \caption{\textit{Disease module for classification of kidney cancer survival and non-survival patients. Left panel: Disease subnetwork with importance of the edges ($Imp_{e}$), indicated by edge thickness. Right panel: Multi-modal importance measures of the involved omics ($IMP_{f}$).}}
\end{figure*}

\begin{table}[ht]
\centering
\caption{Performance of the detected module}
\begin{tabular}{l c  l  c  c  c}
  \hline
 Experiment & Sensitivity & Specificity & Recall  & Precision & Accuracy \\
  \hline			
  Survival status & 0.79 &  0.44 & 0.79 & 0.66 & 0.62 \\
  Cancer type & 0.74 &  0.79 & 0.74 & 0.79 & 0.76 \\
  \hline  
\end{tabular}
\end{table}

The Decision Tree associated with the top-ranked detected module had a high sensitivity (which is a crucial and important parameter in the medical domain), but the precision of the predictions were moderate. A rather low value for this specific experiment was expected, because the detection of network modules causing a death outcome in patient survival analysis is a hard task. 

We compared the results with a standard RF approach, where we selected the top-15 (same number as for the size of the detected module) ranked features according to the in-build random forest impurity scores. We trained a random forest based on these features, and applied the trained model to the test data set.

While the specificity was slightly higher with $0.5$, the sensitivity was $0.67$, and thus significantly lower as for our module detected approach ($0.79$, see Table 5). These results suggest, that feature selection based on features organized in modules perform superior on unseen test data. In future applications we will include features from a wide range of biological entities, including medical data, while fine-tuning the PPI network, which might further improve the performance.

The genes forming the detected module point to an interesting biological ATP protein complex. ATP proteins are mitochondrial transport proteins and were recently verified as potential biomarker for kidney survival prognosis \cite{trisolini2021differential}. In addition to that, we report on a special importance of the ATP6V1C2 protein for survival prediction (see Figure 6, right panel). Furthermore, the ATP6V1C2-ATP6V0B protein interaction has the highest edge importance score in that module, with $IMP_{e}=0.52$. This finding may suggest the ATP6V0B gene as an important co-expression partner (Figure 6, left panel).  

A potentially important role of ATP6V1C2 is also reported by the tree-based Shapley values based on the test set predictions of the greedy DF classifier (see Figure 7, left panel). DNA Methylation data for that gene does not contribute to the importance values, which can be seen also from Figure 6, and is confirmed in Figure 7, right panel. 

% TreeSHAP 
\begin{figure*}[h!]
  \centering
  \includegraphics[scale=0.46]{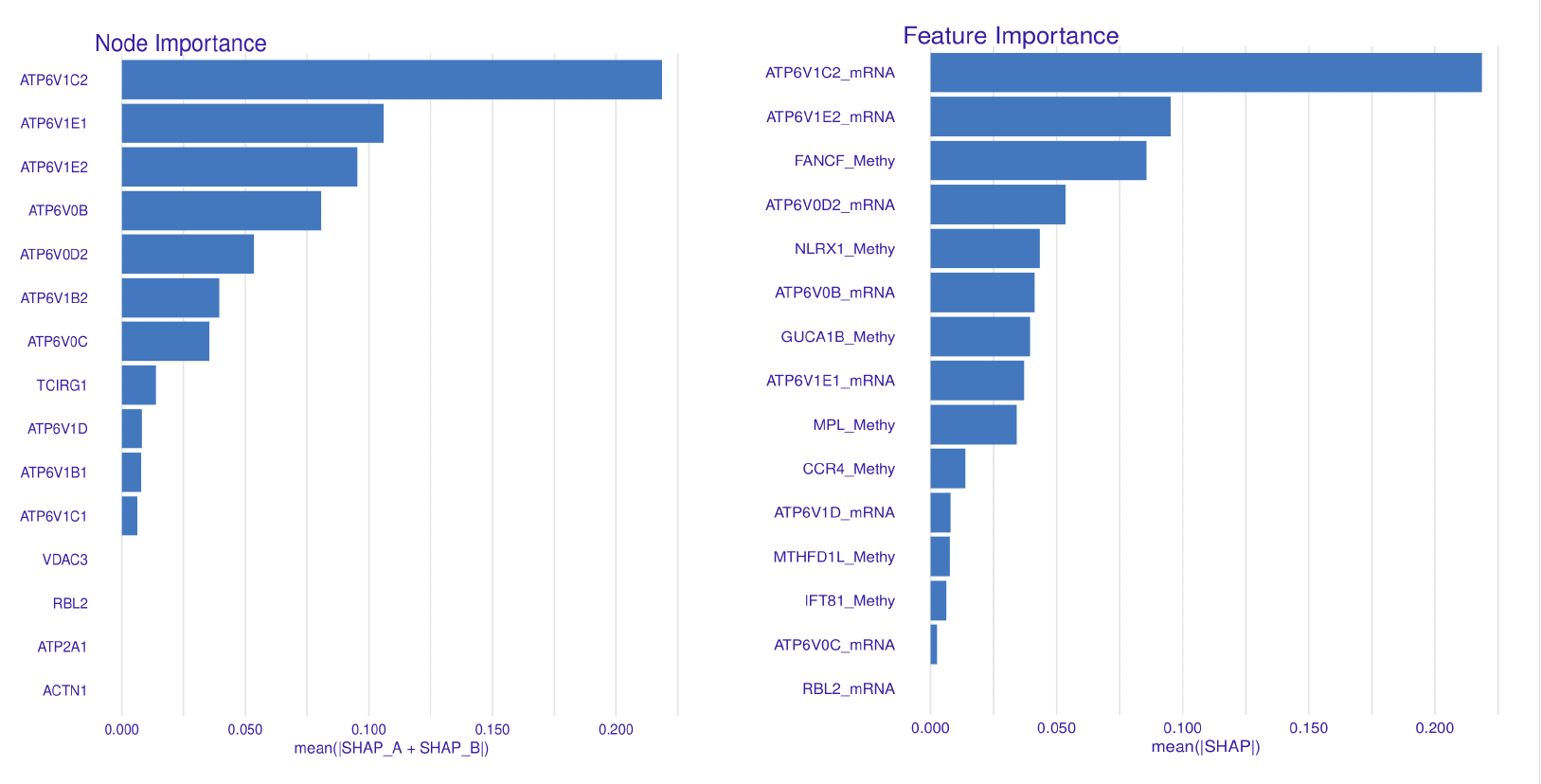}
  \caption{\textit{Tree-based Shapley value explanations on the greedy decision forest classifier}}
\end{figure*}

Node importance by Shapley-values rank the ATP gene family at the very top (see Figure, 7 left panel). From the biological point of view that is a much easier result to interpret, since the ATP genes are functional related, and thus there is a stronger biological evidence for these genes to be important for disease progression. If the decision forest would have been trained on independent features, it is likely that the top-ranked features were widely distributed over the entire PPI network.   

%TreeSHAP for a single patient
\begin{figure*}[h!]
  \centering
  \includegraphics[scale=0.55]{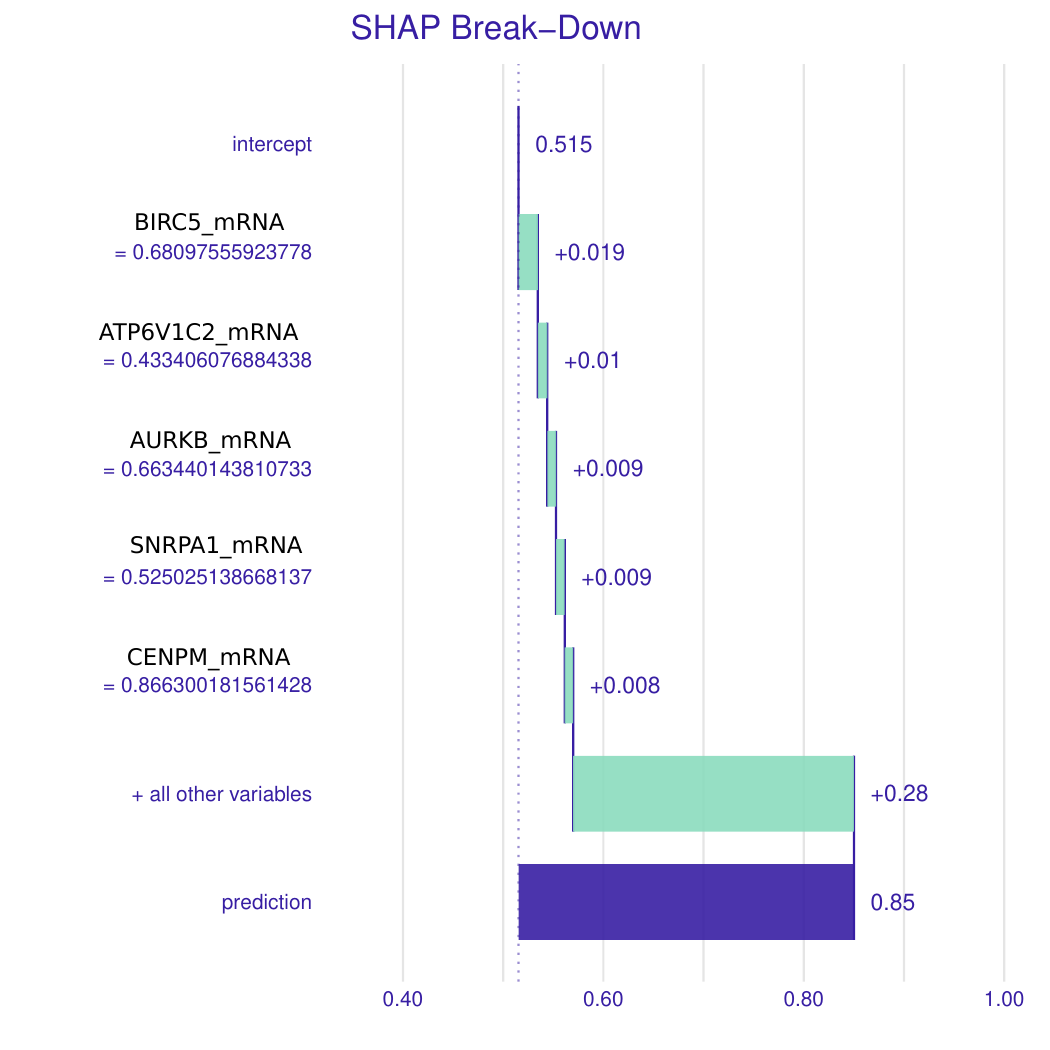}
  \caption{\textit{Explanations on a single non-survived patient with tree-based Shapley values}}
\end{figure*}

With shapley-values it is also possible to explain predictions for a single patient. As an example we report on the feature importance of a non-survived patient (see Figure 8). In this case only a few variables contributed significantly more than the others. Interestingly, all of these features are based on gene expression signatures. ATP6V1C2 is within that set, but does not have a much influence. The reason for that observation might be the high sensitivity of our classifier. High gene expressions within the ATP gene family might contribute to a overall better survival state, but might not be used for a predictor for a death outcome. It can be observed that many more variables are involved in that case (Figure 8). 

Results of the cancer-specific experiment are shown in Table 3 and Table 4. Overall, the predictive accuracy and the quality of the detected modules is higher compared to the more difficult task of finding biomarker modules for disease survival prediction.  

%%%%%%%%%%%%%%%%%%%%%%%%%%%%%%%%%%%
\section{Discussion}

In this work, we have proposed a tree-based network module selection algorithm. It can be applied to the scenario where each node is linked to a specific feature vector and any kind of dependency between these features are reflected by the edges of a given domain knowledge network. We have validated our approach on synthetically generated Barabasi Networks. We could show that our approach can be naturally applied to the case of multi-modal input features, where each node comprises feature vectors from multiple modalities. Also, our method can be applied to multi-graphs, which is an important feature since missing data is a well known bottleneck within the biomedical domain.

We showcase the applicability of our approach on a protein-protein interaction (PPI) networks, where each node is enriched by multi-omics features, to detect disease causing modules. We believe we have developed a useful framework which may find wide applications in the future, in all areas of research, but especially within the bio-medical domain. This is because the greedy decision forest for module selection is a glass-box approach and thereby decisions it makes, or modules it detects, are traceable by design. 
Deep Learning methods on graphs may solve similar tasks in the future, but are black-box models. Explainable AI methods are further needed to uncover the underlying decision path. In fact, our proposed greedy Decision Forest may act as a baseline for further developments in that field solving similar problems by the deep learning community. 

%Also, our methodology might be modified to a explainable GNN method. Instead of increasing the performance of the DF classifier with regard to a target vector, our method may be adjusted to converge to the predicted outcomes of a GNN classifier. The glasss-box greedy DF classifier would serve as a global explainer for black-box models, while the Decision Trees could provide local explanations. This is similar to what we have discussed in \cite{HolzingerEtAl:2021:GraphFusion}, and will be part of future work.       

However, several technical improvements to the proposed algorithm could be made. %First, the mechanism to explore the network structure can be more sophisticated. Here, we have adopted a standard random walk algorithm applicable on unweighted networks, as implemented within the R-package igraph. In future work, we plan to employ a random walk which can be applied also to weighted graphs. 
The PPI network retrieved from the STRING database contains information about the certainty about a functional relationship between two proteins. It is reflected by a "combined score" value, which may change depending on future biological experiments and research findings. Therefore, a weighted random walk could be utilized to explore paths within the PPI network using edge specific certainty scores.  
Second, the $niter$ (number of greedy steps) is a crucial parameter. We not yet provide a stopping criteria. It is similar to the problem of choosing the right number of trees within a Random Forest. We plan to work on a mechanism which monitors convergence during the greedy runs while improving the sampling strategy to accelerate exploration.

Furthermore, we aim for testing the greedy Decision Forest using other than the AUC-ROC performance measure \cite{Zheng:2018:FeatureEngineering}, or even a combination thereof by voting. Because of possible complications with the AUC-ROC on unbalanced data sets \cite{CarringtonEtAl:2020:AUC}, we advise to apply our algorithm on balanced data sets, e.g accommodated by up, or down sampling.     
%Finally, the glass-box nature of this algorithm enables actionable explainability; this is not a necessary condition, since there are other methods that aid actions, like Layer-wise Relevance Propagation \cite{Yeom:2021:PruningByExplaining}. Nevertheless, it is a sufficient one, since it guides the domain expert and the AI system's designer in a transparent way towards actively improving the dataset the model, and the explainable method itself. This is going to be a major topic in future experiments that use this algorithm.

%%%%%%%%%%%%%%%%%%%%%%%%%%%%%%%%%%%%%%%

\section{Conclusion}

We propose a novel approach for selecting disease modules from multi-omic node feature spaces while including domain-knowledge into the algorithmic pipeline. Our approach can be applied to a variety of possible applications where network structured data form the input and the corresponding nodes are associated with multi-modal numerical feature values. Disease module detection in PPI networks is a typical example of such an application. This work represents a contribution to the field of explainable AI for systems biology. Our proposed greedy decision forest is straightforward to interpret. All decision paths are explainable and interpretable by a human domain expert and any contribution of features from different modalities can be retrieved. Tree-based Shapley values can be utilized for explanations, when our greedy decision forest is utilized as a machine learning classifier. We could showcase, that Shapley values report on biological more relevant patterns when the tree model is learned on PPI networks. 

In future work, we will extend our framework for explicit patient survival analyses, taking into account survival status as well as survival times (e.g. using tree-based risk prediction models). Furthermore, the analysis of mixed input data will be supported, whereby both categorical and numerical node features of arbitrary dimensions can be considered. While we believe that our method is very useful for systems biology applications, this method is also applicable in other domains, where the underlying data can be structured as a graph and nodes are enriched by feature values.

%We propose a novel approach for network module selection from multi-modal node feature spaces. Our approach can be applied to a wide range of possible applications, where network structured data build the input, and corresponding nodes are associated with multi-modal numerical feature values. The detection of disease modules in PPI networks is a typical example for such an application.
%Our greedy Decision Forest is easy to interpret. All decision paths are traceable by a human domain expert, and the contribution of the features from the modalities can be easily accessed.  
%In future work, we will extend our framework for explicit patient survival analyses, taking into account survival times in addition to the survival status (e.g through tree-based risk prediction models). Furthermore, the analyses of mixed input data will be supported, allowing for categorical as well as for numerical node features of arbitrary dimensions. 
%%%%%%%%%%%%%%%%%%%%%%%%%%%%%%%%%%%%%%

\section*{Code availability}
\noindent We have implemented our approach within the R-package DFNET, freely available on GitHub (https://github.com/pievos101/DFNET). DFNET interfaces with the R-package ranger \cite{Rranger}. Networks are organized and visualized using the R-package igraph \cite{Rigraph}. Tree-based Shapley values are calculated using the R-package treeshap \cite{Rtreeshap}.

\section*{Abbreviations}

\begin{itemize}

\item AXAI = Actionable Explainable Artificial Intelligence
\item DT = Decision Tree
\item DF = Decision Forest
\item RF = Random Forest
\item GDF = Greedy Decision Forest
\item FS = Feature Selection
\item NN = Neural Network
\item GNN = Graph Neural Network
\item AUC = Area Under the Curve
\item ROC = Receiver Operating Characteristic
\item BN  = Barabasi Network
\item PPI = Protein-Protein Interaction
\item ML = Machine Learning
\item DL = Deep Learning 
\item SHAP = SHapley Additive exPlanation
\item XAI = Explainable Artificial Intelligence

\end{itemize}

%%%%%%%%%%%%%%%%%%%%%%%%%%%%%%%%%%%%%%

\section*{Acknowledgments}

\noindent Parts of this work have received funding from the European Union's Horizon 2020 research and innovation programme under grant agreement No.826078 (Feature Cloud). This publication reflects only the authors' view and the European Commission is not responsible for any use that may be made of the information it contains. Parts of this work have been funded by the Austrian Science Fund (FWF), Project: P-32554 ``explainable Artificial Intelligence''. 

%%%%%%%%%%%%%%%%%%%%%%%%%%%%%%%%%%%%%%%%%%%%%%%%%%%%%%%%%%%%%%%%%%%%%%%%%%%%%%%%%%%%%%%
\bibliography{references}

\end{document}